\documentclass[10pt, a4paper]{article}
\usepackage{graphicx} 
\usepackage[margin=3.25cm]{geometry}
\usepackage{mlmodern}
\usepackage{amsfonts}
\usepackage{amsmath}
\usepackage{setspace}
\usepackage[numbers]{natbib}
\usepackage{float}
\usepackage{dsfont}
\usepackage{hyperref}
\usepackage{enumerate}
\usepackage{enumitem}
\usepackage{subcaption}
\usepackage{multirow}
\usepackage{multicol}
\usepackage{lineno}
\usepackage[ruled,vlined, linesnumbered]{algorithm2e}

\newlength\mylen
\newcommand\myinput[1]{%
  \settowidth\mylen{\KwIn{}}%
  \setlength\hangindent{\mylen}%
  \hspace*{\mylen}#1\\}
\let\oldnl\nl
\newcommand{\nlnonumber}{\renewcommand{\nl}{\let\nl\oldnl}}

\begin{document}

\begin{center}
{\Large
	{\sc Uncertainty Guarantees on Automated Precision Weeding using Conformal Prediction\footnote{Preprint. Currently under review.}}
}
\bigskip

 \underline{Paul Melki} $^{1, 2}$, Lionel Bombrun $^{1,3}$, 
 Boubacar Diallo $^{2}$,\\ Jérôme Dias $^{2}$ \& Jean-Pierre Da Costa $^{1,3}$
\bigskip

{\it
$^{1}$ Univ. Bordeaux, CNRS, Bordeaux INP, IMS, UMR 5218, F-33400 Talence, France \\
$^{2}$ EXXACT Robotics, F-51200 Épernay, France \\
$^{3}$ Bordeaux Sciences Agro, F-33175 Gradignan, France \\
\texttt{paul.melki@exxact-robotics.com}
}
\end{center}
\bigskip

\begin{abstract}
    \noindent Precision agriculture in general, and precision weeding in particular, have greatly benefited from the major advancements in deep learning and computer vision. A large variety of commercial robotic solutions are already available and deployed. However, the adoption by farmers of such solutions is still low for many reasons, an important one being the lack of trust in these systems. This is in great part due to the opaqueness and complexity of deep neural networks and the manufacturers' inability to provide valid guarantees on their performance. Conformal prediction, a well-established methodology in the machine learning community, is an efficient and reliable strategy for providing trustworthy guarantees on the predictions of any black-box model under very minimal constraints. Bridging the gap between the safe machine learning and precision agriculture communities, this article showcases conformal prediction in action on the task of precision weeding through deep learning-based image classification. After a detailed presentation of the conformal prediction methodology and the development of a precision spraying pipeline based on a ``conformalized'' neural network and well-defined spraying decision rules, the article evaluates this pipeline on two real-world scenarios: one under \textit{in-distribution} conditions, the other reflecting a \textit{near out-of-distribution} setting. The results show that we are able to provide formal, \textit{i.e.} certifiable, guarantees on spraying at least $90\%$ of the weeds.
\end{abstract}


\section{Introduction}
Precision agriculture, just like a multitude of other science and engineering domains, has greatly benefited from the large and quick-paced advancements in machine learning over the last decade \citep{condranMachineLearningPrecision2022, liakosMachineLearningAgriculture2018}. The methods have been ingeniously adapted – and sometimes improved – to applications in agriculture, spanning plant phenotyping \citep{gillComprehensiveReviewHigh2022, solimaniSystematicReviewEffective2023}, disease identification \citep{dominguesMachineLearningDetection2022}, weed detection \citep{colemanWeedDetectionWeed2022, hasanSurveyDeepLearning2021}, yield estimation \citep{paudelInterpretabilityDeepLearning2023, palaciosEarlyYieldPrediction2023}, and soil monitoring \citep{vermeulenMachineLearningPerformance2017, patriziVirtualSoilMoisture2022}, among others. These methods have allowed practitioners to keep pace with the ever-increasing quantity and diversity of data associated with precision agriculture practices. Indeed, multiple data modalities are now involved, often in the same system or use case. These include RGB, multispectral and hyperspectral images, text, LiDAR and RADAR data, GNSS and location data, and acoustic signals to name only a few \citep{kamilarisReviewPracticeBig2017}.

Deep neural networks are singled out, in both scientific literature and industrial applications, as the most powerful approach to “make sense” of this considerable flood of data \citep{kamilarisDeepLearningAgriculture2018a}. These multi-layered mathematical models whose architecture can be adapted to the type of data in input and the desired output have become more common with the advancements of software tools that allow users to import, train and deploy them with relatively simple programming steps \cite{chollet2015keras, paszkeAutomaticDifferentiationPyTorch2017}. Their adoption has also been facilitated by the advances in adapted hardware and their wider availability (such as graphical processing units) along with the development of tools and methods that have allowed their embedding and deployment on the edge \textit{in the wild} for various agricultural applications \citep{duLightweightDeepNeural2022, yuNeuralNetworksRealtime2023}.

Although neural networks have shown their superiority in terms of predictive accuracy on a majority of applications and benchmark datasets, they still lag behind other predictive models on a number of highly desirable characteristics \citep{daraRecommendationsEthicalResponsible2022}, such as good interpretability \citep{rudinInterpretableMachineLearning2022, zhangSurveyNeuralNetwork2021}, robustness to aberrant observations \citep{madryDeepLearningModels2018, raghunathanUnderstandingMitigatingTradeoff2020}, and good estimation of prediction confidence \citep{guoOnCalibration2017, niculescumizilPredictingGoodProbabilities2005, ovadiaCanYouTrust2019}. The opaqueness and complexity of neural networks, combined with an often ignored caveat – which is the lack of valid guarantees on the quality of the predictions under real-world conditions \citep{vovkAlgorithmicLearningRandom2005, damourUnderspecificationPresentsChallenges2022, hendrycksUnsolvedProblemsML2022}, puts the users of deep learning models in front of a difficult conflict between predictive performance ``in the lab'', and a certain lack of trust in the predictive system to be deployed in the real-world. This trade-off is not only faced by developers of precision agriculture systems employing deep learning approaches. It has repercussions on the end-users of the systems, the farmers. Scientists and manufacturers propose systems that not only contain vital components that the farmers do not understand, but also to which we do not associate trustworthy guarantees. This gets translated as a lack of trust in the systems proposed to the practitioners and leads, in consequence, to a reduced adoption of precision agriculture systems \citep{alexanderWhoResponsibleResponsible2023}. Farmers are reluctant to invest in technologies that do not come with quality guarantees because the economic and environmental risks associated with system malfunction are high \citep{pathakSystematicLiteratureReview2019, tzachorResponsibleArtificialIntelligence2022}.

Conformal prediction \citep{vovkAlgorithmicLearningRandom2005, angelopoulosConformalPredictionGentle2023a, shaferTutorialConformalPrediction2008, papadopoulosInductiveConfidenceMachines2002a} has gained much attention in recent years in the machine learning community. This methodology is an efficient and reliable strategy for providing formal valid guarantees on the predictions of any black-box model, including neural networks, under very minimal constraints. This makes it very attractive for real-world deployment. It constitutes an important step towards shipping trustworthy and certifiable machine learning models \citep{huangSurveySafetyTrustworthiness2020, balasubramanianConformalPredictionReliable2014}. While conformal prediction has already been applied in a number of fields where safety and reliability are critical,  such as medical diagnosis \citep{papadopoulosReliableDiagnosisAcute2009, luFairConformalPredictors2022}, autonomous driving \citep{lindemannSafePlanningDynamic2023a}, robotics \citep{ lekeufackConformalDecisionTheory2024, luoSampleefficientSafetyAssurances2024}, and even nuclear fusion \citep{vegaAccurateReliableImage2010}, among others \citep{balasubramanianConformalPredictionReliable2014, andeolConfidentObjectDetection2023, vilfroyConformalPredictionRegression2024}, its integration into agricultural and environmental use cases is still limited and in its early stages \citep{chiranjeeviDeepLearningPowered2023, faragInductiveConformalPrediction2023b, kakhaniUncertaintyQuantificationSoil2024, melkiGroupConditionalConformalPrediction2023c, melkiPenalizedInverseProbability2024a}. 

The aim of this article is to bridge the gap between the safe machine learning and precision agriculture communities by showcasing conformal prediction in action on the task of precision weed spraying through deep learning-based image classification. Working with a proprietary database of weed and crop images acquired in Europe, we develop a conformal prediction-based pipeline that \textit{guarantees} the detection of at least $90\%$ of weeds. \\

\noindent \textbf{\textit{Main Contributions --}} The article's main contributions may summarized as follows:

\begin{itemize}
    \item A detailed presentation of the conformal prediction methodology to the ag-tech community, including an overview of multiple nonconformity scores, two conformal algorithms (marginal split-conformal prediction and its class-conditional version) and the evaluation metrics;
    \item The development of a conformal prediction-based precision spraying pipeline using a ``conformalized'' neural network image classifier and well-defined spraying decision rules;
    \item The evaluation of this pipeline and its comparison to a standard classification model under two experimental procedures reflecting real-world scenarios: one under \textit{in-distribution} conditions, the other simulating a \textit{near out-of-distribution} setting.
\end{itemize}

In Section \ref{sec:data} we begin by presenting the extensive database collected internally and used to conduct the study. Section \ref{sec:conformal-prediction} constitutes the foundation of this work. It begins with a general introduction to conformal prediction and its main ingredients followed by a presentation of the central notion of \textit{nonconformity score} and a detailed description of multiple scores studied here. We then present two important algorithms: split-conformal prediction and class-conditional conformal prediction. Section \ref{sec:base-classifier} quickly presents the neural network used in the pipeline. Section \ref{sec:decision-rules} details the rules defined in the pipeline for deciding to spray or not, and Section \ref{sec:eval-metrics} presents the various evaluation metrics used in our work. In Section \ref{sec:exper-proc} we present, the data used in the empirical study and, in turn, each experiment: its setup and then its results. In Section \ref{sec:discussion} we discuss the experimental results and the role of conformal prediction in this context, then conclude in Section \ref{sec:conclusion}.

\section{Materials and Methods}

\subsection{Conformal Prediction} \label{sec:conformal-prediction}
\subsubsection{Overview and Main Components}
\label{sec:overview_cp}
\begin{figure}[ht]
    \centering
    \includegraphics[width=0.85\linewidth]{ 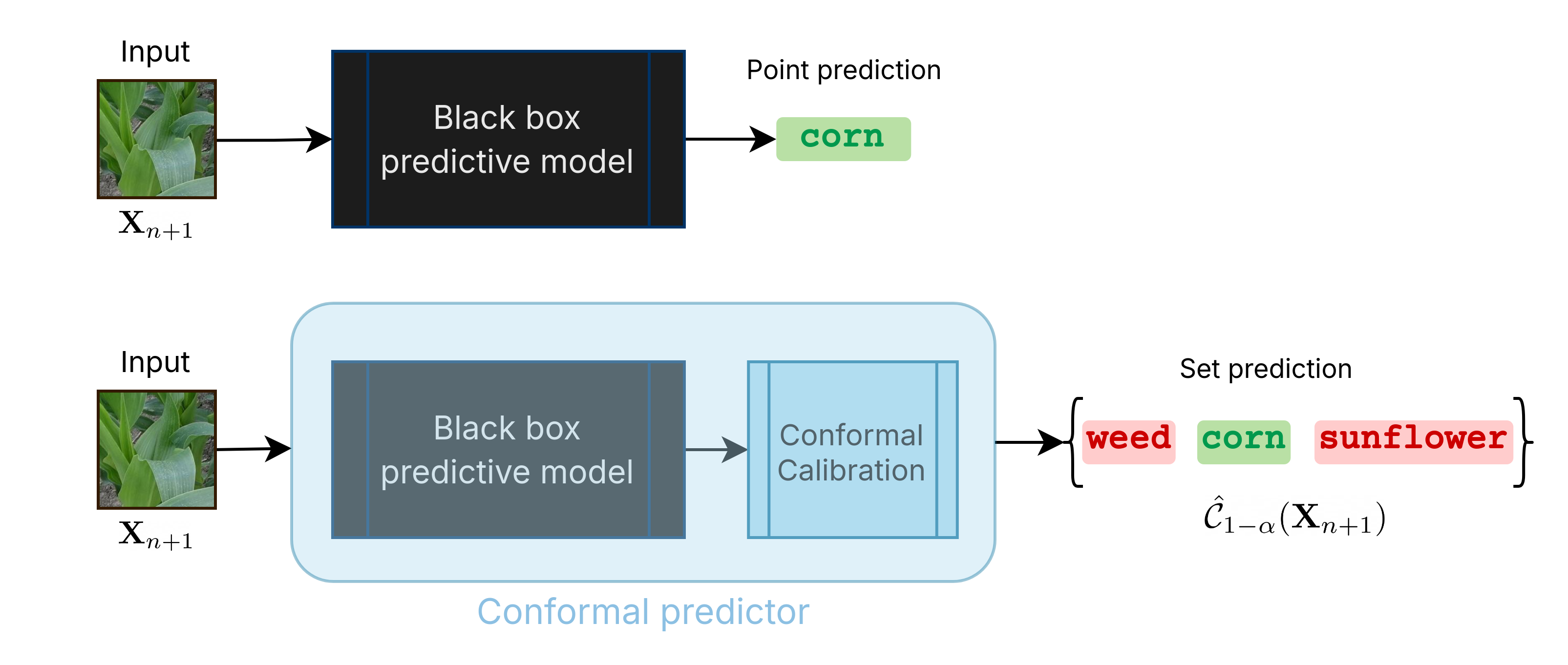}
    \caption{A high level representation of how conformal prediction transforms any black-box point predictor into a conformal set predictor.}
    \label{fig:cp_high_level}
\end{figure}
Conformal Prediction \citep{vovkAlgorithmicLearningRandom2005} is a general framework for quantifying and controlling uncertainty in machine learning models without explicit assumptions on the data distribution. The essential idea is quite simple: it consists of transforming any predictive model that produces point predictions into a set or interval predictor (Figure \ref{fig:cp_high_level}). The prediction sets are constructed in such a way as to guarantee the inclusion of the true value with a probability of $1-\alpha$, where $\alpha \in (0, 1)$ is the tolerance chosen by the user, typically taken to be 0.05 or 0.1. Concretely, assume we have a couple of random variables $(\mathbf{X}, Y) \in \mathcal{X} \times \mathcal{Y}$ from which we observe $n$ examples  $\{ (\mathbf{x}_1, y_1), ... ,(\mathbf{x}_n, y_n) \} $. For instance, $\mathbf{X}$ can be an image and $Y$ its associated ground-truth class, as in our use case. For a new observation $\mathbf{X}_{n+1}$, the goal of a conformal predictor calibrated on the $n$ examples is to produce a prediction set, $\hat{\mathcal{C}}_{1-\alpha}(\mathbf{X}_{n+1}) \subseteq \mathcal{Y}$, satisfying the following \textit{marginal coverage} property:
\begin{equation}
    \mathbb{P} \big(Y_{n+1} \in \hat{\mathcal{C}}_{1-\alpha}(\mathbf{X}_{n+1}) \big) \ge 1 - \alpha
\label{eq:marginal_coverage}
\end{equation}
This guarantee means that \textit{on average over all observations}, the predicted set of $Y$ values for a new observation will contain the true value $(1 - \alpha) \times 100\%$ of the times. It thus allows the user to control the error rate of the predictive system and to associate formal guarantees to its predictions, which is not the case for most of the ordinarily used point predictors without conformal calibration, including neural networks. This guarantee is valid under the relatively weak assumption of the exchangeability of the data \citep{aldousExchangeabilityRelatedTopics1985, vovkAlgorithmicLearningRandom2005}.

As a general framework, conformal prediction has been adapted to a large spectrum of learning tasks (classification, regression, segmentation, etc.), predictive models (SVM, decision trees, neural networks, etc.), and data configurations (offline vs. online learning) \citep{fontanaConformalPredictionUnified2023, angelopoulosConformalPredictionGentle2023a}. While each application of conformal prediction may have its specificity, all conformal approaches share the same general requisites:

\begin{itemize}
    \item The user's \textit{tolerance to error}, $\alpha$. This fixed parameter, to be chosen by the user before the start of the conformal procedure is what defines the significance level $1- \alpha$ at which the coverage guarantee is to be maintained. Based on this level, the conformal pipeline will then ensure that the ground-truth label is included no less than $(1- \alpha) \times 100 \;\%$ of the times in the predicted sets as per Equation \ref{eq:marginal_coverage}. This parameter is usually defined based on the use case's requirements.

    \item The \textit{base predictor}, hereby noted as $\mathcal{B}$, which can be any black-box point predictor on which to apply the conformal procedure. In this article, we work exclusively with neural network classifiers.

    \item Following the \textit{split-conformal} procedure \citep{papadopoulosInductiveConfidenceMachines2002a} (see Section \ref{sec:split_conformal}) requires three distinct non-overlapping datasets. Namely, a first \textit{training} set of $n_{\text{train}}$ observations on which $\mathcal{B}$ is trained. The conformal calibration procedure, which will transform $\mathcal{B}$ into a conformal set predictor requires a separate set of $n_{\text{cal}}$ \textit{calibration} observations, and, finally, a set of $n_{\text{test}}$ observations on which to evaluate the performance. 

    \item Perhaps the most important component, the \textit{nonconformity score} $\Delta$, can be understood as a ``distance" that measures how similar -- or \textit{conforming} -- a new observation $\mathbf{X}_{n+1}$ is to the previously seen data. Different scores have previously been introduced in the literature and will be discussed in detail in the next section.
\end{itemize}

\subsubsection{Nonconformity Score Functions}
\label{sec:nc_scores}

Before we review the different nonconformity scores for classification, it is important to define some mathematical notations required for the rest of the article.

As mentioned previously, we assume that we observe a random vector $\mathbf{x} \in \mathcal{X}$ of features, called an \textit{object}. Each object is assigned a ground-truth class label $y \in \mathcal{Y} := \{1,..., K\}$ by the annotators. The couple $(\mathbf{X}, Y) \in \mathcal{X} \times \mathcal{Y}$ is often called an \textit{example}. The classifier $\mathcal{B}$ can be any function that predicts a class $\mathcal{B}(\mathbf{x}) = \hat{y} \in \mathcal{Y}$ for an input $\mathbf{x} \in \mathcal{X}$ after it has been trained on $n_{\text{train}}$ examples. We also assume that $\mathcal{B}$ also produces some sort of associated estimated probability $\hat{p}^{\hat{y}} \in [0, 1]$, such that $\sum_{k=1}^K \hat{p}^k = 1$. The neural network classifier used in this study is a typical example of such a setup, since for each class we can associate its corresponding softmax output which can be interpreted as a heuristic estimation of probability \citep{guoOnCalibration2017}.

The nonconformity score $\Delta(\mathbf{z}): \mathcal{X} \times \mathcal{Y} \rightarrow \mathbb{R}$ is a real-valued function that associates to each example a measure of ``strangeness'' with respect to the previously seen examples. It is a crucial component of both the conformal calibration and conformal prediction procedures, and, in a certain sense, is what distinguishes one conformal predictor from another. Here, we will cover some of the nonconformity scores that are especially suitable for the task of classification using neural networks \citep{johanssonModelAgnosticNonconformityFunctions2017, angelopoulosUncertaintySetsImage2021, melkiPenalizedInverseProbability2024a}. To simplify the notation, we write the nonconformity score as a function of the label $y$ since the object $\mathbf{x}$ is implicitly included in the output of the base classifier.

\begin{itemize}
    \item \textbf{Hinge Loss (IP)} \citep{johanssonModelAgnosticNonconformityFunctions2017}. A very ``natural" score of nonconformity is the \textit{Inverse Probability},  i.e. \textit{Hinge Loss}:
\begin{equation} \label{eq:hinge}
        \Delta^{\text{IP}}(y) = 1 - \hat{p}^y \;.
\end{equation}
While very intuitive and easy to compute, this score has a major deficiency in that it is blind to the estimated probabilities of other classes than the class of interest.

\item \textbf{Margin Score (MS)} \citep{johanssonModelAgnosticNonconformityFunctions2017}. The \textit{Margin Score} measures how different the estimated probability of the class of interest $y$ from is the highest estimated probability among the other classes:
\begin{equation} \label{eq:margin}
    \Delta^{\text{MS}}(y) = \max_{k \ne y} \hat{p}^k - \hat{p}^y \; .
\end{equation}
This score assumes that when class $y$ has a large margin away from the class with the highest estimated probability, it should be considered ``nonconformal". Such a hypothesis is implicitly based on the idea that the true class should often be the one with the highest probability.

\item \textbf{Adaptive Prediction Sets (APS)}  \citep{romanoClassificationValidAdaptive2020, angelopoulosUncertaintySetsImage2021}. The APS conformal method has been introduced with the aim of predicting sets whose size is adaptive to the perceived difficulty of the input object by the base classifier. Unlike the two previously described scores, APS does actually take into consideration the estimated probabilities of all the classes deemed ``more probable" than the class of interest $y$, following this definition:
\begin{equation} \label{eq:raps}
    \Delta^{\text{APS}}(y) = \sum_{r = 1}^{R(y)-1} \hat{p}^{[r]} + u \cdot \hat{p}^{[R(y)]} \; ,
\end{equation}
where $R(k)$ is the rank of class $k$ after the classes have been sorted by decreasing order of estimated probabilities, $\hat{p}^{[r]}$ is the probability estimate of the class having rank $r$, such that $\hat{p}^k = \hat{p}^{[R(k)]}$, and $u$ is a random value sampled uniformly in $(0, 1)$ for breaking potential ties \cite{angelopoulosUncertaintySetsImage2021}.

\item \textbf{Penalized Inverse Probability (PIP)} \citep{melkiPenalizedInverseProbability2024a}. Our recently proposed nonconformity score introduces a penalization term to the classical IP score. This penalization takes into consideration the classes having higher estimated probabilities than the class of interest:
\begin{equation} \label{eq:pip}
    \Delta^{\text{PIP}}(y) = 1 - \hat{p}^y + \sum_{r = 1}^{R(y)-1} \frac{\hat{p}^{[r]}}{r} ~\mathds{1}_{\{R(y) > 1 \}} \; ,
\end{equation}
where the penalization is only added when $y$ is the not the most probable class, where in such cases the score is simply the Hinge Loss (IP). This score has been introduced with the aim of minimizing the average size of the predicted sets while maximizing the number of singletons, which is a non-trivial optimization problem that is not fulfilled by the other nonconformity scores.
\end{itemize}

A detailed discussion of the attributes and behavior of these different nonconformity scores under different conditions can be found in our previous work \citep{melkiPenalizedInverseProbability2024a}. In this article, these different scores will be tested in our experimental procedure with the aim of choosing the most suitable one for the precision spraying pipeline.

\subsubsection{Split-Conformal Procedure}
\label{sec:split_conformal}

\begin{algorithm}
\caption{Split-Conformal Procedure} \label{algo:split_conf}

\BlankLine
\SetKwInOut{Input}{Input}\SetKwInOut{Output}{output}
\SetKwFunction{Append}{Append}
\KwIn{\textit{calibration set} $\{ \mathbf{z}_i = (\mathbf{x}_i, y_i), i = 1,..., n_{\text{cal}}\}$}
\nlnonumber \myinput{\textit{test set} $\{ \mathbf{z}_i = (\mathbf{x}_j, y_j), j = 1,..., n_{\text{test}}\}$}
\nlnonumber \myinput{\textit{error tolerance} $\alpha \in (0, 1)$}
\nlnonumber \myinput{\textit{trained base model} $\mathcal{B}$}
\nlnonumber \myinput{\textit{nonconformity score function} $\Delta(\;)$}
\nlnonumber \myinput{\textit{number of classes} $K$}

\KwOut{prediction sets $\left\{\hat{\mathcal{C}}_{1-\alpha}(\mathbf{x}_j), j = 1,..., n_{\text{test}} \right\}$}

\BlankLine

Train base model $\mathcal{B}$ on training set \;
\nlnonumber \textsc{Conformal Calibration} \\ 
\For{$i \leftarrow 1$ \KwTo $n_{\text{cal}}$}{
Predict using trained model: $\mathcal{B}(\mathbf{x_i})$\;
Use prediction to compute nonconformity score of true class: $\Delta(y_i)$\;
}
Estimate $q_{\text{cal}}$, the $1-\alpha$ quantile of $\left\{\Delta(y_i), i=1,...,n_{\text{cal}} \right\}$\;

\nlnonumber \textsc{Conformal Prediction} \\
\For{$j \leftarrow 1$ \KwTo $n_{\text{test}}$}{
Predict using trained model: $\mathcal{B}(\mathbf{x_j})$\;
$\mathcal{\hat{C}}_{1-\alpha}(\mathbf{x_j}) \leftarrow \{ \; \}$\;
\For{$k \leftarrow 1$ \KwTo $K$} {
Compute nonconformity score of considered class: $\Delta(k)_j$\;
\If{$\Delta(k)_j \leq q_{\text{cal}}$}{
    $\mathcal{\hat{C}}_{1-\alpha}(\mathbf{x_j}) \leftarrow \mathcal{\hat{C}}_{1-\alpha}(\mathbf{x_j}) \cup k$
    }
}
}

\end{algorithm}

The \textit{split-conformal} approach is a simple and efficient procedure to conduct conformal prediction in an offline setting \citep{papadopoulosInductiveConfidenceMachines2002a}. It consists of two main steps: (1) conformal calibration, and (2) conformal prediction (Algorithm \ref{algo:split_conf}).

As described in Section \ref{sec:overview_cp}, this procedure requires three distinct non-overlapping datasets. The training set is used for training the base model $\mathcal{B}$. After the training phase, the base model does not change anymore. That is, the intrinsic parameters that have been learned are not modified by the additional conformal component added to the predictive pipeline. The calibration dataset, is used for the conformal calibration step that will be shortly described, and the test set is used for testing the conformal model.

The conformal calibration step is essentially the step that transforms $\mathcal{B}$ which produces point predictions into a set predictor (Figure \ref{fig:cp_high_level}). For each example $\mathbf{z}_i = (\mathbf{x}_i, y_i), i = 1,...,n_{\text{cal}}$ in the calibration set, we compute the nonconformity score of the true class $y_i$. We thus obtain the set of ``true'' nonconformity scores on the calibration set: $\left\{ \Delta(y_i), i = 1,...,n_{\text{cal}} \right\}$. We then estimate the $1-\alpha$ quantile value of these scores, $q_{\text{cal}}$, where $\alpha$ is the chosen level of tolerance to error. This quantile value is the main output of the conformal calibration procedure and will be used at the prediction phase.

Indeed, $q_{\text{cal}}$, being the $1 - \alpha$ quantile, is the value after which the $\alpha$ most ``extreme'' nonconformity scores lie. The examples that lie after $q_{\text{cal}}$ are the $10\%$ examples with highest nonconformity scores. That is, the most ``nonconforming'' observations. 

Following this reasoning, it becomes easier to understand how conformal prediction works. It can be seen as a form of hypothesis test whereby we test how ``conforming'' a certain class is relatively to the calibration data. Indeed, for a given new input $\mathbf{x}_j, j \in [1, n_{\text{test}}]$ from the test set, we compute the probability estimates $\hat{p}_j^k$ for each class $k = 1,..., K$ by passing the input in the base model. For each class, we compute its nonconformity score $\Delta(k)_j$ and compare it with the calibration quantile $q_{\text{cal}}$. If the $\Delta(k)_j$ lies in the rejection region, it is deemed too extreme with regards to the previously seen data. Class $k$ is therefore considered highly unlikely to be the true class, and is thus excluded. Otherwise, $k$ is considered plausible enough to be the true class and is therefore included in the prediction set. It is important to note that in the predicted sets, the classes included are sorted in increasing order of nonconformity scores.

Following this procedure, under the crucial hypothesis that the calibration and test data come from the same -- or at least similar -- population, provably leads to conformal sets that satisfy the desired marginal coverage property defined in Equation \ref{eq:marginal_coverage}. We can thus guarantee that the true class is covered by our prediction sets \textit{at least} $(1 - \alpha) \times 100\%$ of the times \citep{papadopoulosInductiveConfidenceMachines2002a, vovkAlgorithmicLearningRandom2005}. 

\subsubsection{Class-Conditional Procedure}
\label{sec:cc_proc}
The split-conformal procedure provides only a \textit{marginal} guarantee on the coverage of the true class. This means that, on average over all examples, the coverage is guaranteed at $1 - \alpha$. However, this does not mean that the coverage is maintained for any arbitrary example $(\mathbf{x}, y)$, a guarantee proven to be impossible to achieve with a conformal predictor with bounded average set size \citep{foygelbarberLimitsDistributionfreeConditional2021}. Furthermore, the marginal coverage condition does not guarantee that the coverage is maintained at $1 - \alpha$ for each possible sub-group of the examples. In particular, the coverage may not be guaranteed for each class \cite{cauchoisKnowingWhatYou2021}. This can have dire consequences in sensitive applications where the coverage has to be maintained equally for all classes or on a particular class of interest. For example, in the current use case of precision weeding, it is much more important to provide valid guarantees on class \textit{weed} which we aim to eradicate than on the other \textit{background} or crop classes.

Formally, we would like to achieve the \textit{class-conditional coverage} guarantee:
\begin{equation}
    \mathbb{P} \big(Y_{n+1} \in \hat{\mathcal{C}}_{1-\alpha}(X_{n+1}) | Y_{n+1} = y \big) \ge 1 - \alpha \;\;\; \forall y \in \mathcal{Y}
\end{equation}
which can be satisfied by iteratively applying the conformal calibration procedure separately on the examples belonging to each class in the calibration set. That is, for each $y \in \mathcal{Y}$, we conduct conformal calibration on the examples $(X_i, y), i = 1,..., n_{\text{cal}}$ to obtain a class-conditional quantile $q^{(y)}_{\text{cal}}$ for each class. At prediction time, the nonconformity score of each class $y$ is tested against its respective quantile $q^{(y)}_{\text{cal}}$ to decide whether to include $y$ in the prediction set or not. This procedure provably guarantees the class-conditional coverage \citep{vovkConditionalValidityInductive2013, cauchoisKnowingWhatYou2021, angelopoulosConformalPredictionGentle2023a}.

\subsection{Base Classifier} \label{sec:base-classifier} 
As mentioned previously, conformal prediction is an added layer on top of a base classifier $\mathcal{B}$. As conformal prediction is a model-agnostic approach, any classifier that produces some form of probability estimate for each class is a good candidate for becoming a conformal predictor, even if it does not manifest high classification performance \citep{romanoClassificationValidAdaptive2020}. As such, a classical ResNet-50 \citep{heDeepResidualLearning2016} classifier is considered in this study, although more recent and performing models can be used \citep{schusterConsistentAcceleratedInference2021}.

\subsection{Decision Rules on Conformal Sets}
\label{sec:decision-rules}
The prediction sets allow us to provide reliable guarantees on the performance of the predictive model. However, when integrated in a pipeline where a final decision or action should be taken, it is important to define the decision functions that map the prediction sets into decisions \citep{straitouriImprovingExpertPredictions2023, lekeufackConformalDecisionTheory2024}. 

\begin{figure}[ht]
    \centering
    \includegraphics[width=\linewidth]{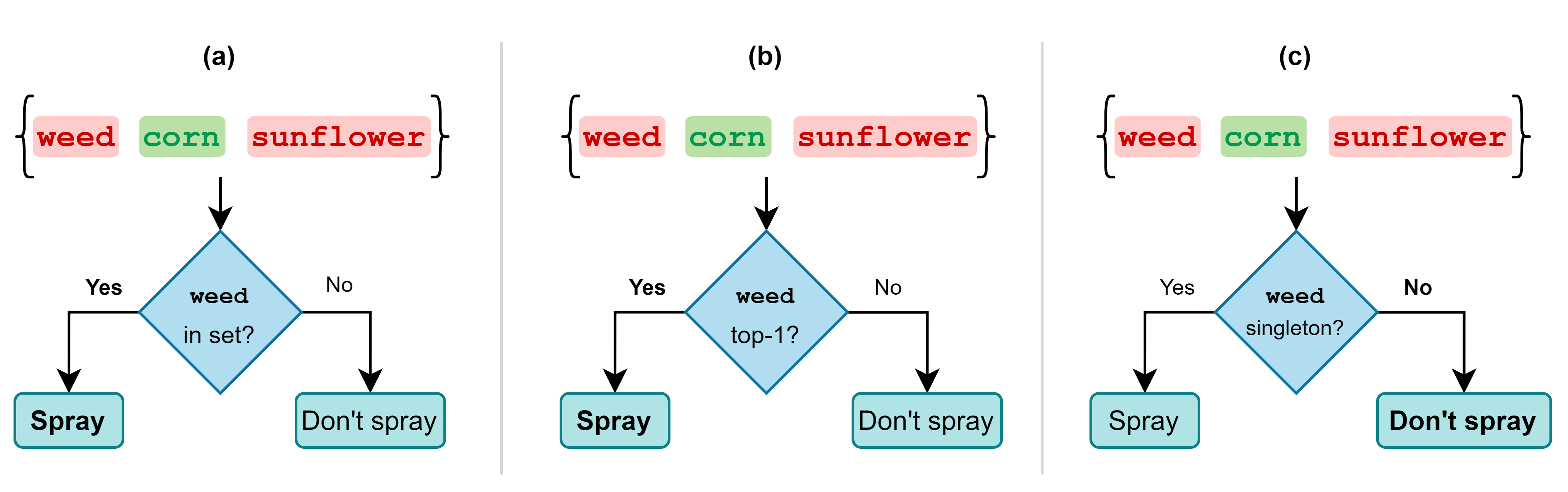}
    \caption{Schematic representations of the three spraying decision functions considered in our spraying pipeline. The decision taken in each example is shown in bold. (a) \textit{Weed in set}, (b) \textit{Weed top-1}, (c) \textit{Weed singleton}. Classes are ranked by decreasing order of $\hat{p}^k$, \textit{corn} in green is the true class label.}
    \label{fig:decision_rules}
\end{figure}

The precision spraying pipeline that we develop has to output a final binary \texttt{spray} / \texttt{no spray} decision. As such, we define three possible binary in top of the predicted sets. All of these functions rely on the existence of class \textit{weed} in the obtained sets (Figure \ref{fig:decision_rules}).

\begin{enumerate}[label=(\alph*)]
    \item \textit{Weed in set:} The first decision function leads to a \texttt{spray} decision for a certain input, whenever the class \textit{weed} is included in the predicted set. This can be considered as the simplest and most lenient decision function and therefore may potentially lead to spraying in unneeded locations. For example, in Figure \ref{fig:decision_rules}(a), a spraying decision is taken, even though the true class label is \textit{corn}, simply because \textit{weed} has also been predicted as a plausible class. 

    \item \textit{Weed top-1:} As mentioned previously, the predicted classes are sorted in decreasing order of nonconformity scores in the sets, from the class considered the most ``conforming", to the least plausible class. Accordingly, we can define a decision function where \texttt{spray} is decided only when class \textit{weed} is the first predicted class (Figure \ref{fig:decision_rules}(b)). This decision function is more stringent than the previous one in that it only sprays when \textit{weed} is the most plausible class.

    \item \textit{Weed singleton:} When a singleton is predicted, it means that the one class predicted has been considered overwhelmingly more conforming than the other classes: the base model manifests very high certainty about the class it predicts. The strictest decision function is thus the one where the model chooses to spray when \textit{weed} is the only predicted class (Figure \ref{fig:decision_rules}(c)).
\end{enumerate}

\subsection{Evaluation Metrics}
\label{sec:eval-metrics}
There are three levels of evaluation in the precision spraying pipeline considered in this work. The first level consists of evaluating the base model on the multiclass classification task after training it. For this, we use standard classification metrics such as \textit{precision}, \textit{recall}, and their harmonic mean, the \textit{f1-score} \cite{grandiniMetricsMultiClassClassification2020}. This is not the level of evaluation that interests us since the rest of the conformal pipeline provides guaranteed coverage regardless of the performance of the base model \citep{johanssonModelAgnosticNonconformityFunctions2017}. \\

\noindent \textbf{\textit{Evaluating the conformal predictor}} \quad The second level evaluates the performance of the conformal predictor. In order to verify that the coverage guarantee defined in Equation \ref{eq:marginal_coverage} is maintained, we can compute the \textit{empirical marginal coverage}, which is simply the average number of times the true class has been covered by the predicted set: 
\begin{equation} \label{eq:emp_marginal_coverage}
    \text{Empirical Coverage} = \frac{1}{n_{\text{test}}} \sum_{i=1}^{n_{\text{test}}}  \mathds{1}_{\{ y_i \in \mathcal{C}_{1 - \alpha}(\mathbf{x}_i) \}}
\end{equation} 
where $\mathds{1}_{\{\cdot\}}$ is the indicator function. This value should be around the $1-\alpha$ level of coverage.

While all conformal predictors are valid by construction, they are not all equally ``useful"~\citep{babbarUtilityPredictionSets2022, straitouriImprovingExpertPredictions2023}. Indeed, the usefulness of a set predictor can be measured using two metrics introduced in the literature \citep{vovkCriteriaEfficiencyConformal2016, johanssonModelAgnosticNonconformityFunctions2017}. 

\begin{itemize}
    \item The \textit{efficiency} is simply defined as the average size of the predicted sets:
    \begin{equation} \label{eq:efficiency}
        \text{Efficiency} = \frac{1}{n_{\text{test}}} \sum_{i=1}^{n_{\text{test}}} | \mathcal{C}_{1 - \alpha}(\mathbf{x}_i) |
    \end{equation} 
    where $|~\cdot~|$ is the set cardinality, the number of classes in the predicted set. Clearly, smaller sets are preferred since they are easier to interpret and to construct decision rules on. 

    \item The \textit{informativeness} computes the ratio of singleton predictions:
    \begin{equation} \label{eq:informativeness}
        \text{Informativenesss} = \frac{1}{n_{\text{test}}} \sum_{i=1}^{n_{\text{test}}}  \mathds{1}_{\{| \mathcal{C}_{1 - \alpha}(\mathbf{x}_i)|  = 1\}}
    \end{equation}
    Singleton predictions are the most efficient and the easiest to interpret. Therefore, a conformal predictor that predicts more singletons without violating the coverage guarantee is preferred.
\end{itemize}

A conformal model that jointly optimizes these two criteria can be readily used in a decision-making pipeline as the one proposed in the current work. \\

\noindent \textbf{\textit{Evaluating the spraying decision}} \quad The third level of evaluation is that of the spraying decisions: are we taking the right spraying decision when weeds are detected? To estimate the spraying quality, we rely on two main important assumptions related to the data (\textit{cf.} Section \ref{sec:data}). Namely, that each image is well-aligned with a spraying head and that it represents one unit of spraying surface. We also consider that each crop or weed is seen only once and that no chemical spray drift occurs \citep{partelDevelopmentEvaluationLowcost2019}. In other words, each image, and in consequence each spraying decision, is associated to one plant. While these assumptions may be considered simplifying ones, they are crucial to allow a reliable estimation of sprayed ratio in an \textit{a posteriori} experimental setup where a physical inspection of the land is not possible. We argue that given the large amount of data available, such an estimation can provide a reliable evaluation of the spraying performance of a given pipeline.

As such, the evaluation at this level consists of comparing the \textit{sprayed ratio}, estimated by the number of spraying decisions over the number of all images in the considered sample. The \textit{sprayed ratio} can be formally defined as: 
\begin{equation}
    \text{Sprayed Ratio} = \frac{1}{n_{\text{test}}} \sum_{i=1}^{n_{\text{test}}} \mathds{1}_{\{\text{``spray"}\}} \; .
\end{equation}
This ratio can be compared to the \textit{infestation level}, estimated as the number of images annotated as \textit{weed} over the number of images in the considered sample:
\begin{equation}
    \text{Infestation Level} = \frac{n_{\text{\textit{weed}}}}{n_{\text{test}}} \; .
\end{equation}
Here, $n_{\text{test}}$ is the number of images in the studied sample. In addition, two other metrics of interest can be inferred from these quantities. The \textit{spray reduction} level with respect to broadcast spraying (where the whole parcel surface is sprayed) which can be computed as
\begin{equation}
    \text{Spray Reduction} = 1 - \text{Sprayed Ratio},
\end{equation}
is a metric that can be of particular interest to end-users (the farmers) since it is directly related to the reduction of production costs. This reduction may be offset by the \textit{spray surplus}, which is the unneeded amount sprayed, estimated as the excess of spraying decisions taken with respect to the infestation level:
\begin{equation}
    \text{Spray Surplus} = \text{Sprayed Ratio} - \text{Infestation Level}
\end{equation}
This quantity can be negative when not enough spraying decisions were taken, indicating clearly that some weeds have been missed.

\section{Experimental Procedures \& Results} \label{sec:exper-proc}

\subsection{Precision Spraying Image Dataset} \label{sec:data}
To empirically study the conformal prediction-based spraying pipeline, we constructed a large dataset of 510,600 RGB images of dimensions $224 \times 224$ where each image is associated to a potential spraying location. That is, an image is assumed to be well-aligned with a spraying head and represents a spraying surface.

Working in the context of non-selective precision weeding, where the weed species need not be identified, each image is carefully annotated by agronomists who associate to it one ground-truth class among six possible classes: the four studied crops -- \textit{corn}, \textit{rapeseed}, \textit{sugar beet} and \textit{sunflower} --, \textit{weed} whenever a weed  or undesired plant is found (even if another crop is apparent in the image), and finally \textit{background} when no plant appears in the image. 

\begin{figure}[ht]
    \centering
    \includegraphics[width=0.7\linewidth]{ 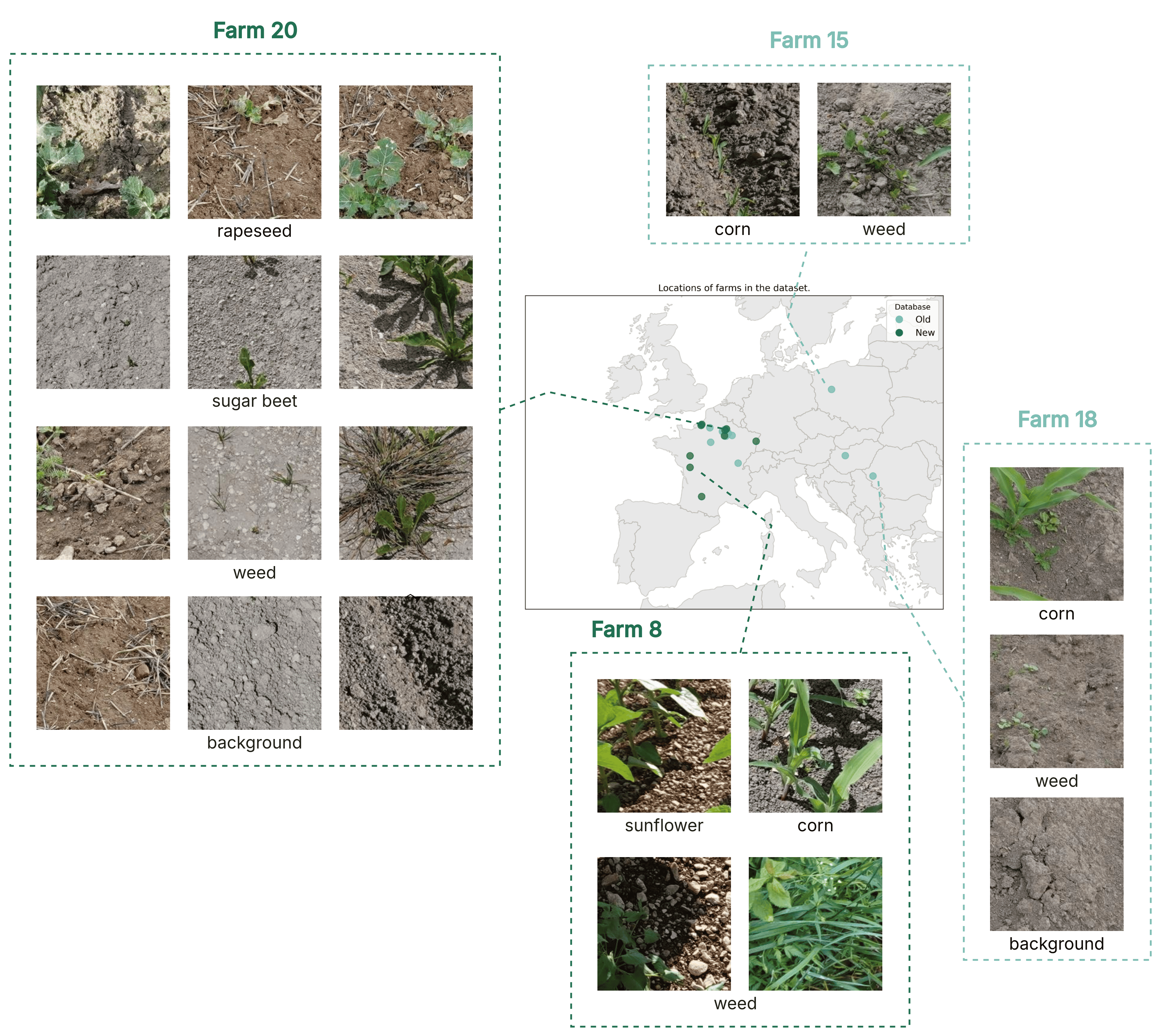}
    \caption{Example images and their ground-truth classes showing the diversity of conditions across some farms in Europe.}
    \label{fig:example_images}
\end{figure}

Data have been collected in 21 ``farms'' across France and other European countries, manifesting a large diversity of visual and semantic differences, as shown in Figure \ref{fig:example_images}. The collection procedure was conducted in two stages, hence the distinction between the ``old'' and ``new'' databases (Figure \ref{fig:example_images}), an information that will be exploited in the experimental procedure. 

All farms are infested with weeds of different types and so exhibit at least three classes: \textit{weed}, \textit{background} and one or many crop classes. Indeed, some farms contain multiple plots where different crops are cultivated. Only one of the farms, \textit{Farm 20}, contains all six classes.

\subsection{Experiment 1: Typical Lab Conditions}
\subsubsection{Setup}

The first experiment aims at showing the conformal prediction procedure in action under typical ``lab'' conditions. Indeed, the validity of the conformal approach relies on the fundamental hypothesis that the test distribution is the same,  or at least \textit{similar}, to the distribution of the calibration set \citep{vovkAlgorithmicLearningRandom2005}. This hypothesis may not always be satisfied in real-world deployment due to naturally occurring distribution shifts from changes in the visual characteristics of images, in the environmental conditions, or the semantics of the objects of interest \citep{gibbsAdaptiveConformalInference2021, quinonerocandelaDatasetShiftMachine2022, cappioborlinoAddressingDistributionalShift2024}.

In order to guarantee the satisfaction of this hypothesis, we consider in this experiment the \textit{old} database exclusively, and rely on repeated random subsampling of the data, which on average would reflect the distribution of the original database \citep{lohrSamplingDesignAnalysis2019}.

\begin{figure}[ht]
    \centering
    \includegraphics[width=0.7\linewidth]{ 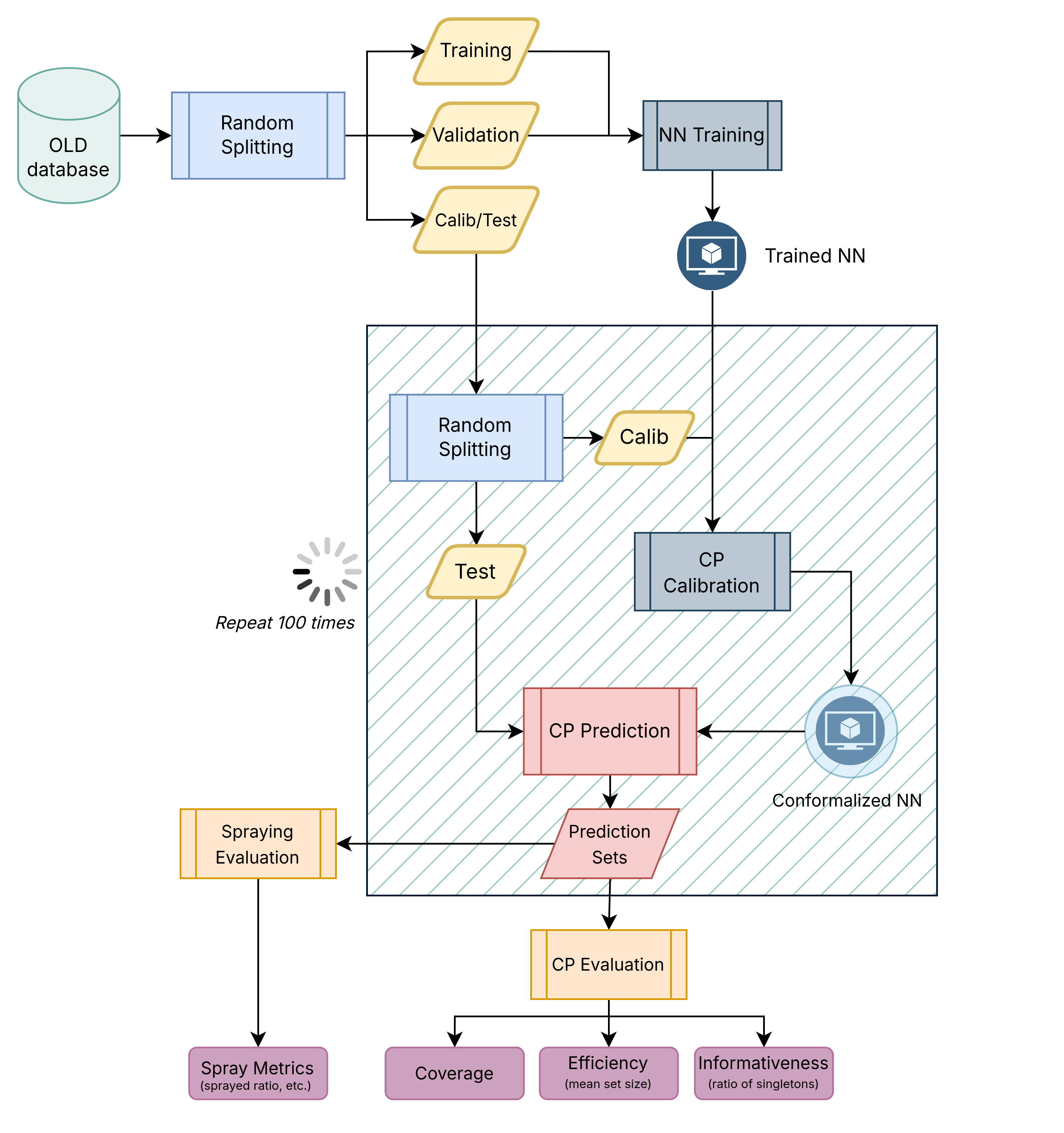}
    \caption{Diagram summarizing the \textit{Experiment 1} procedure. Only the ``old'' database is considered with the aim of showing the behavior of the conformal procedure when the calibration and training sets originate from the same population.} 
    \label{fig:exp1_schema}
\end{figure}

The experimental procedure is detailed in Figure \ref{fig:exp1_schema}. The \textit{old} database is first randomly split into three sets: the training ($50\%$), the validation ($20\%$) and the calibration/test ($30\%$) sets. The training and the validation sets are used to finetune the ResNet-50 neural network pretrained on ImageNet \citep{krizhevskyImageNetClassificationDeep2012} for the task of image classification into \textit{weed}, \textit{background} and crop species classes. After the training phase, the base classifier is fixed and does not undergo any changes. The calibration/test set is then randomly split into the respective calibration set ($45\%$) used for the conformal calibration procedure, and the test set ($55\%$) on which the conformal predictor is tested. The conformal procedure is repeated on 100 different splits of the calibration and test data. The obtained prediction sets are then analyzed in terms of empirical coverage, efficiency and informativeness (defined in Section \ref{sec:eval-metrics}) and then in terms of spraying performance after applying the decision rules on the sets as described in Section \ref{sec:decision-rules}. This procedure is then repeated for each of the nonconformity score functions introduced in Section \ref{sec:nc_scores} with the aim of choosing the optimal score for this use case. The repetition of the conformal procedure across random data splits and nonconformity scores allows us to study the stability of the procedure with regards to the data and the choice of the score function.

\subsubsection{Results}
We take a first look at the classification results of the ResNet-50 models after training (Table \ref{tab:exp1-classif-metrics}). The model has good overall performance across all the classes thanks to the controlled environment in which the experiment is conducted where the training and test data are subsets of the same database. The classifier manifests good recall on our class of interest, \textit{weed}, but is not close to the coverage level that we would like to have at $90\%$, thus showing the incapacity of a coverage control using classical point prediction models. 

\begin{table}[H]
    \centering
    \begin{tabular}{c|c|c|c}
         \textbf{Class} & \textbf{Recall} & \textbf{Precision} & \textbf{F1-Score} \\
         \hline
         \textit{background} & 0.94 & 0.85 & 0.90 \\
         \textit{corn} & 0.82 & 0.91 & 0.86 \\
         \textit{rapeseed} & 0.85 & 0.78 & 0.81 \\
         \textit{sugar beet} & 0.82 & 0.91 & 0.87 \\
         \textit{sunflower} & 0.78 & 0.91 & 0.84 \\
         \textit{\textbf{weed}} & \textbf{0.82} & \textbf{0.74} & \textbf{0.78} \\ 
    \end{tabular}
    \caption{Class-specific classification results of the ResNet-50 after finetuning on the training set. The results are computed on the full calib/test set. The required $1-\alpha = 0.90$ coverage level is not guaranteed for any of the classes using point classification.}
    \label{tab:exp1-classif-metrics}
\end{table}

\begin{figure}[h]
    \centering
    \makebox[\linewidth][c]{
    \begin{subfigure}[b]{0.46\textwidth}
        \centering
        \includegraphics[width=\textwidth]{ 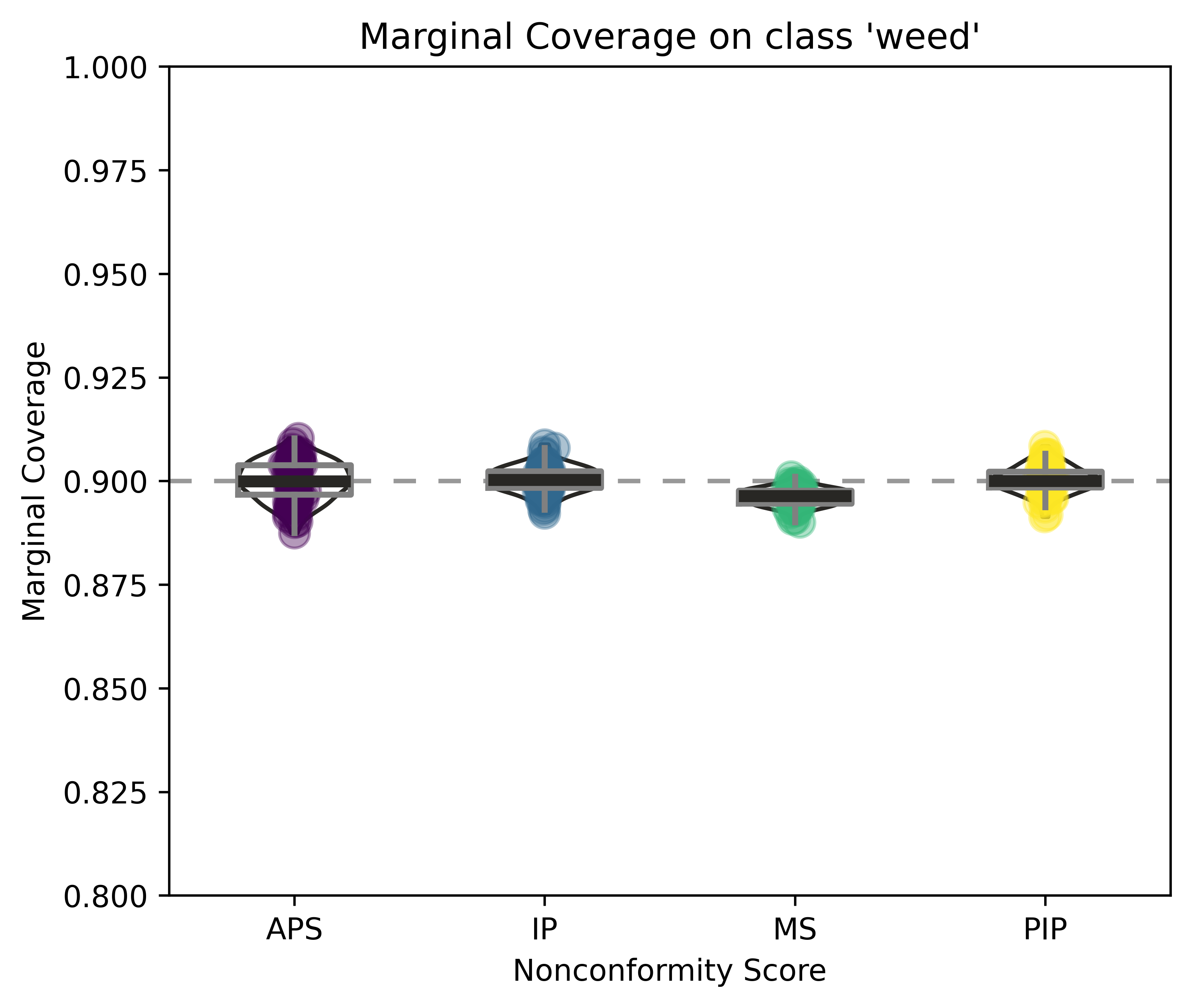}
        \caption{}
        \label{fig:exp1_coverage}
    \end{subfigure}
    \hfill
    \begin{subfigure}[b]{0.45\textwidth}
        \centering
        \includegraphics[width=\textwidth]{ 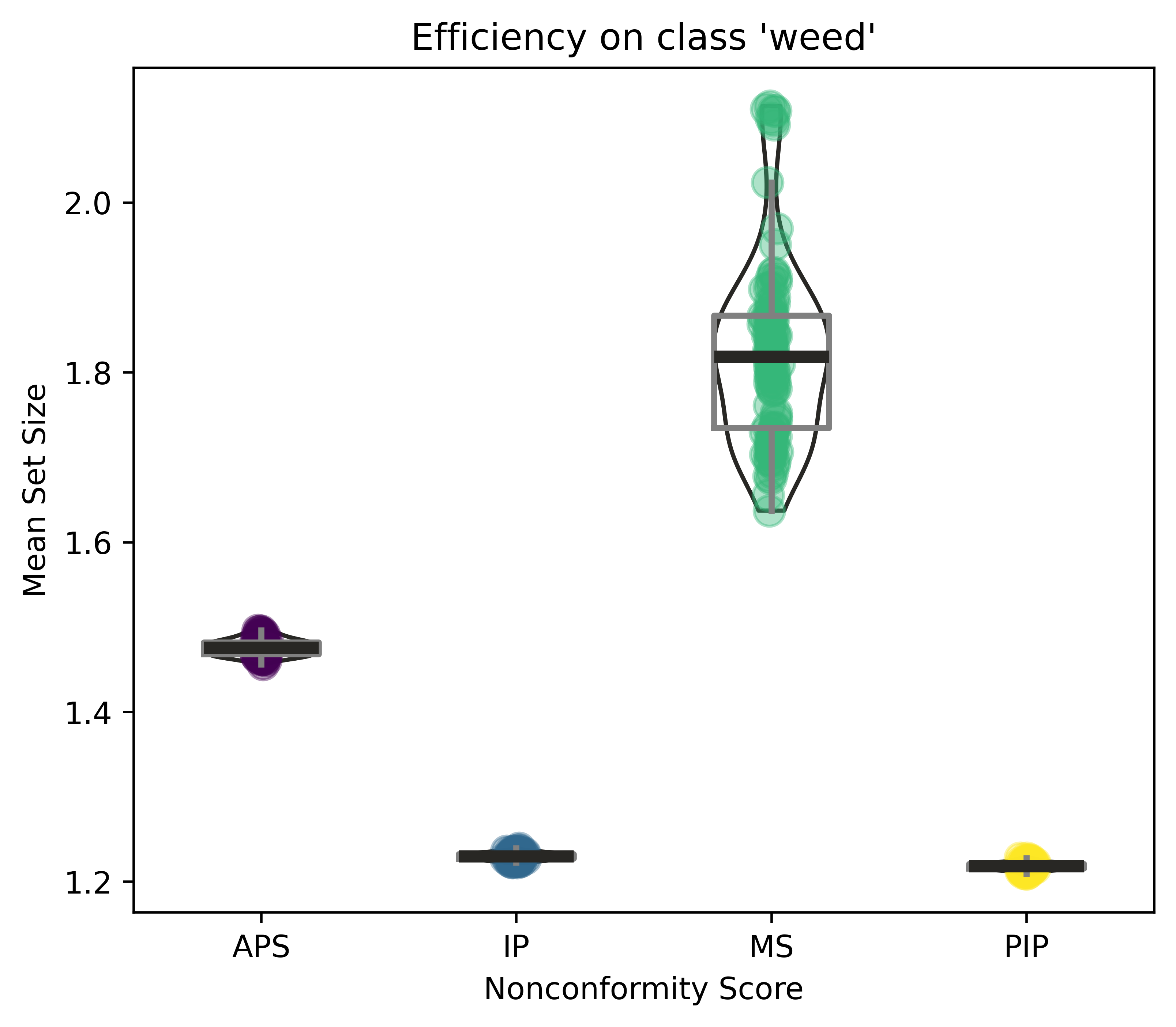}
        \caption{}
        \label{fig:exp1_efficiency}
    \end{subfigure}
    \hfill
    \begin{subfigure}[b]{0.45\textwidth}
        \centering
        \includegraphics[width=\textwidth]{ 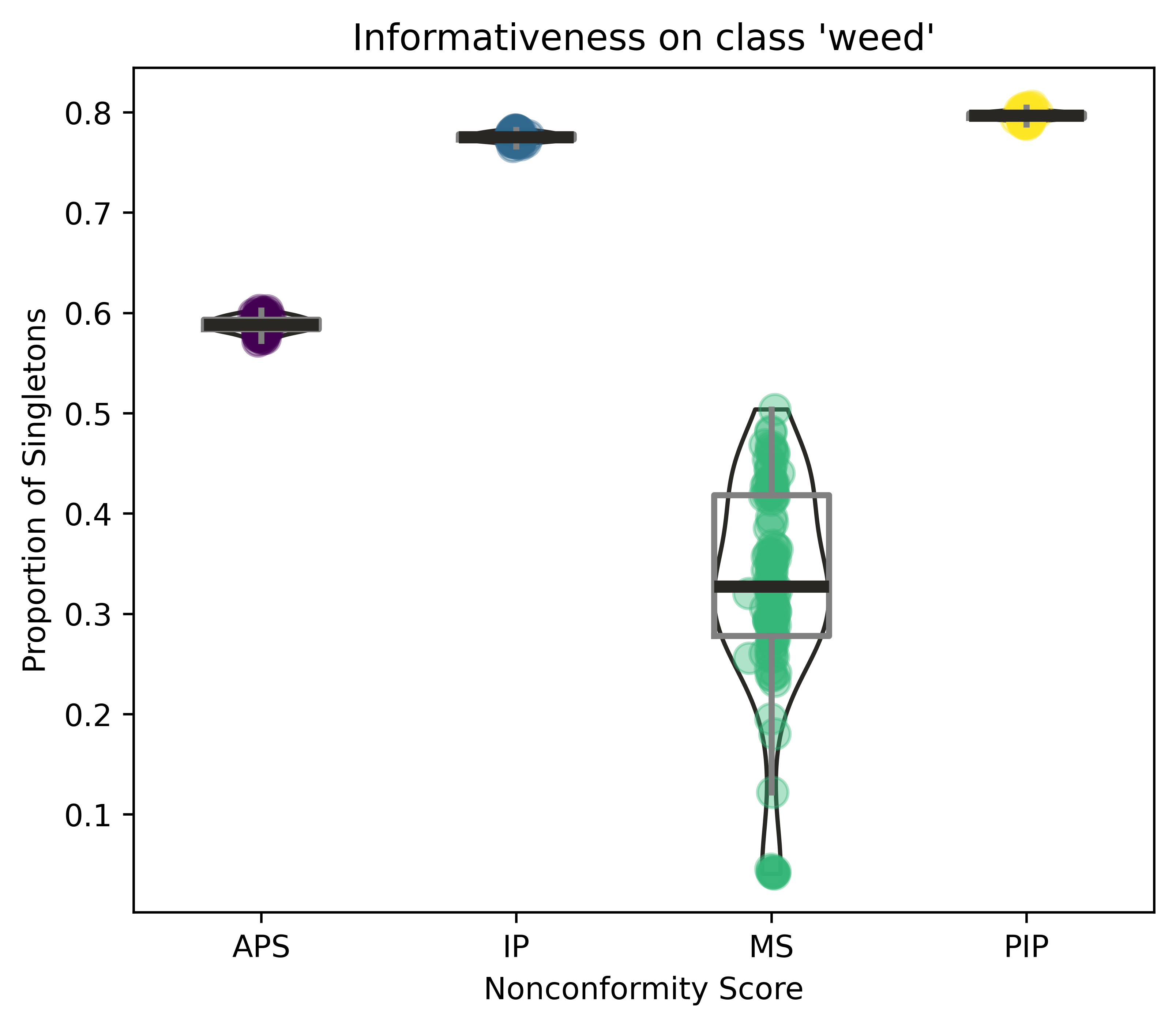}
        \caption{}
        \label{fig:exp1_informativeness}
    \end{subfigure}
    }
    \caption{Conformal evaluation results on class \textit{weed} for the different nonconformity scores over 100 different data splits. (a) Empirical coverage, (b) Efficiency (mean set size), (c) Informativeness (proportion of singletons). While the coverage is maintained on average using all scores, IP and PIP scores are the best in terms of efficiency and informativeness.}
    \label{fig:exp1_conf_results}
\end{figure}

After ``conformalizing'' the base model following the group-conditional calibration procedure, the model is now able to guarantee the coverage on all classes, including class \textit{weed}, as can be seen in Figure \ref{fig:exp1_coverage}. Indeed, as the calibration/prediction procedure is repeated across 100 different subsets coming from the same population, it is not surprising to observe little variability around the $1-\alpha$ coverage level over the different repetitions. However, even though the coverage is guaranteed using any of the nonconformity scores, the IP and PIP scores manifest significantly better performance in terms of efficiency with an average set between 1 and 2 classes  (Figure \ref{fig:exp1_efficiency}) and informativeness with around $80\%$ of predicted sets being singletons (Figure \ref{fig:exp1_informativeness}). In comparison, APS has a significantly lower number of singletons and greater average set size while MS exhibits highly unstable performance and unsatisfactory results. The difference between IP and PIP is not significant in this experimental setup since the base model's performance on \textit{weed} is already not ``too bad'' (Table \ref{tab:exp1-classif-metrics}), which means that PIP will often behave like IP. 

\begin{table}[h]
    \centering
    \resizebox{\textwidth}{!}{
    \begin{tabular}{r|c|c|c|c|c}
    \hline
    \textbf{Infestation Level} & \multicolumn{5}{c}{\textbf{0.29} $\boldsymbol{(\pm 0.002)}$} \\
    \hline
    & \textbf{Recall} & \textbf{Precision} & Sprayed Ratio & Spray Reduction & Spray Surplus \\
    \hline
     \textbf{\textit{Weed in Set}} & 0.90 $(\pm 0.003)$ & 0.63 $(\pm 0.004)$ &  0.40 $(\pm 0.004)$   & 0.60 & 0.11 \\
     \textbf{\textit{Weed Top-1}} & 0.76 $(\pm 0.004)$ & 0.75 $(\pm 0.004)$ & 0.32 $(\pm 0.002)$ & 0.68 & 0.04 \\
     \textbf{\textit{Weed Singleton}} & 0.70 $(\pm 0.004)$ & 0.83 $(\pm 0.004)$ & 0.25 $(\pm 0.003)$ & 0.76 & -0.04\\
\end{tabular}}
    \caption{Average spraying results over the 100 repetitions using the PIP score for the three decision rules (standard deviation between parenthesis).}
    \label{tab:spraying-metrics}
\end{table}

Table \ref{tab:spraying-metrics} shows the results after applying the spraying decision rules. Here we evaluate the decision rules from the perspective of a classical binary decision problem using the recall and precision and then via the metrics that reflect the end-user's needs. The \textit{Weed in Set} decision rule is a direct application of the class-conditional conformal prediction set on class \textit{weed} and therefore is the only one that guarantees the coverage (recall) at the required level of $90\%$. The two other decision rules are far from capturing $90\%$ of the weeds even though they have significantly higher precision than \textit{Weed in Set}: around $83\%$ for \textit{Weed Singleton} and $75\%$ for \textit{Weed Top-1}. This is not surprising since these two functions are more strict than \textit{Weed in Set} and so will tend to be more precise to the detriment of the recall. Looking at the surface-level metrics, we can see that all decision rules lead to a highly significant reduction in spraying in comparison to broadcast spraying (from $60\%$ to $75.5\%$ reduction). Being the least precise, the \textit{Weed in Set} decision rule has the highest spraying surplus: following this rule will lead to a \texttt{spray} decision whenever is the model is suspicious about the existence of a weed. As such, it will sometimes lead to spraying when it is not needed. \textit{Weed Singleton}, on the other hand, is a demanding decision rule and so leads to a spraying shortage while covering only $70\%$ (recall) of the weeds. \textit{Weed Top-1} achieves a nice balance between these two decision rules but still misses $24\%$ of the weeds.

\subsection{Experiment 2: Real-World Scenario}
\subsubsection{Setup}
This second experimental setup is more closely aligned with a typical real-world scenario. The aim of this experiment is to simulate a common use case whereby, after acquiring an original database used for the training of the neural network, we now seek to deploy it on a newly observed farm -- for example, a new client in a commercial context. It has often been noted in machine learning literature that while neural networks may exhibit high performance ``in the lab'', they quite often fail when deployed under real-world conditions that may not closely reflect the original training conditions \citep{stadelmannDeepLearningWild2018, damourUnderspecificationPresentsChallenges2022, hendrycksUnsolvedProblemsML2022, paleyesChallengesDeployingMachine2023}. One potential solution to mitigate this downside is to re-train or finetune the neural network on data from the new domain. For example, when deploying the precision weeding pipeline for a new client, we could acquire new images from the new farms, annotate them, and use them to finetune the model. However, such an approach would require the acquisition of a large amount of new data, which may not always be possible, and the re-training of the neural network which is a relatively long and costly procedure. 
\begin{figure}[H]
    \centering
    \includegraphics[width=0.7\linewidth]{ 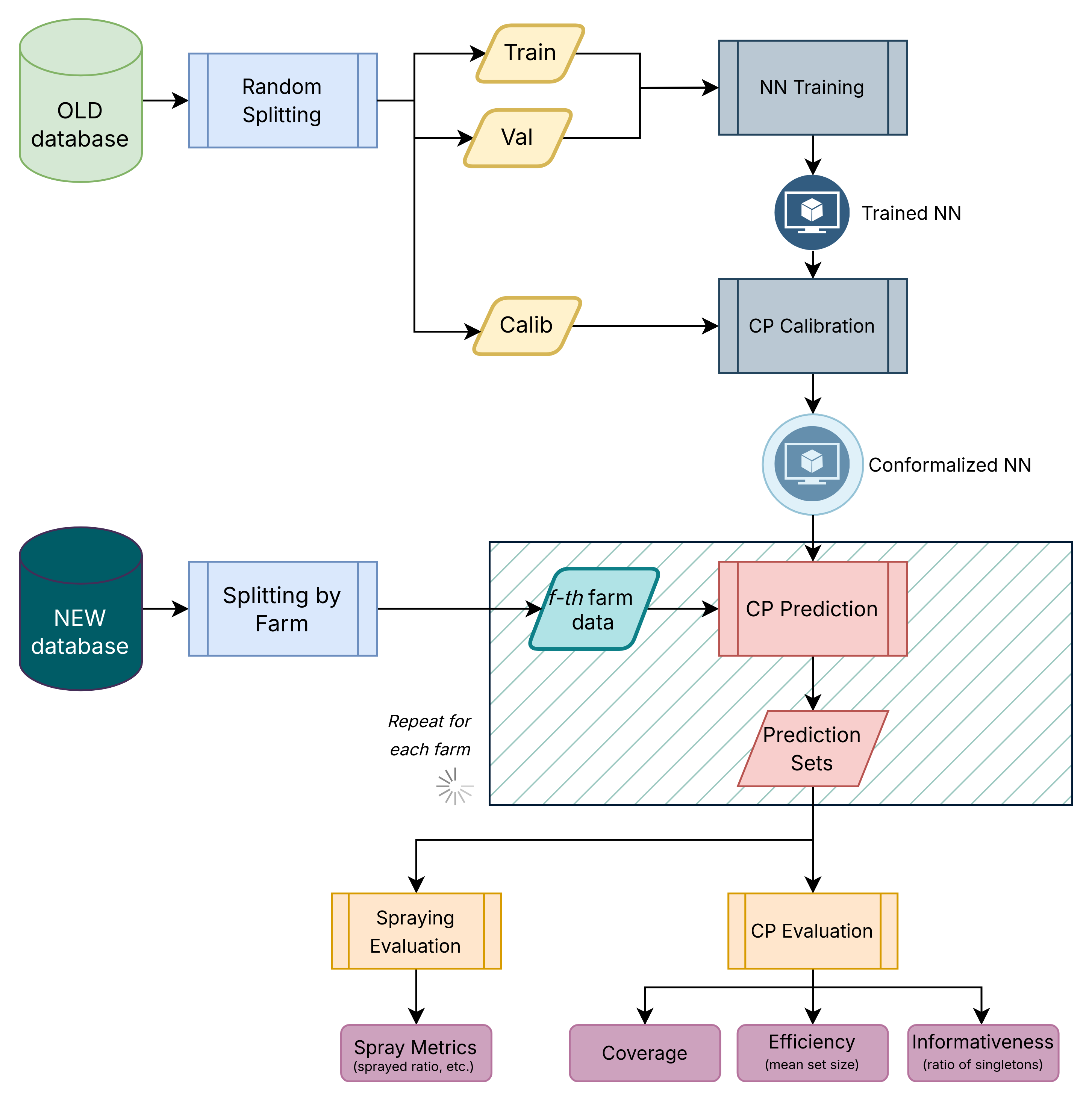}
    \caption{Diagram showing \textit{Experiment 2} procedure. This experimental setup more closely simulates a real-world scenario where a new set of ``farms'' is observed and for which we would like to guarantee the results without re-training the model. } 
    \label{fig:exp2_schema}
\end{figure}

In this experiment, we demonstrate the utility of using conformal predictors in such a scenario. Having a well-trained and calibrated neural network on the \textit{old} database, we empirically show how the model can be readily deployed on \textit{new} farms without violating the coverage guarantee on the class of interest, \textit{weed}. The details of the experimental procedure are shown in Figure \ref{fig:exp2_schema}. Starting with the \textit{old} database, we conduct the same random data split as in Experiment 1 (with the same fixed seed) and conduct the training of the ResNet-50 neural network following the same procedure as the previous experiment. In other words, we use the exact same neural network as previously. The third dataset obtained from the random split is used to calibrate the base model following the class-conditional calibration procedure (Section \ref{sec:cc_proc}). For each farm in the \textit{new} database (Figure \ref{fig:example_images}), we conduct class-conditional conformal prediction. The obtained prediction sets are evaluated using the conformal prediction metrics, and the spraying metrics after applying the decision rules. 

\subsubsection{Results}
Table \ref{tab:exp2-classif-metrics} shows the class-specific classification metrics. There is a clear variability in the model's performance across the different classes, which is probably due to class imbalance in the training data: some classes are significantly more present than others since they can be found in multiple farms. The model is overall less performant than in the previous experiment, indicating that the test population (the \textit{new} database) does not originate from exactly the same population as the \textit{old} database on which the model was trained.
\begin{table}[H]
    \centering
    \begin{tabular}{c|c|c|c}
         \textbf{Class} & \textbf{Recall} & \textbf{Precision} & \textbf{F1-Score} \\
         \hline
         \textit{background} & 0.85 & 0.83 & 0.84 \\
         \textit{corn} & 0.66 & 0.70 & 0.68 \\
         \textit{rapeseed} & 0.45 & 0.76 & 0.57 \\
         \textit{sugar beet} & 0.72 & 0.88 & 0.79 \\
         \textit{sunflower} & 0.51 & 0.68 & 0.58 \\
         \textit{\textbf{weed}} & \textbf{0.82} & \textbf{0.63} & \textbf{0.71} \\ 
    \end{tabular}
    \caption{Class-specific classification results of the ResNet-50 after finetuning on the training set (from the \textit{old} database). The results are computed on the full \textit{new} database.}
    \label{tab:exp2-classif-metrics}
\end{table}

For the conformal procedure, we use the PIP nonconformity score which has shown optimal results in \textit{Experiment 1}. Looking at the obtained results on class \textit{weed} (Table \ref{tab:exp2-conf-results}), we can see that the coverage guarantee is maintained for most of the farms. For some farms, such as \textit{Farm 20} and \textit{Farm 23}, the coverage is far from the required $90\%$ level. This is due to difficulties that these farms present to the model through visual or semantic differences with the calibration data. The model's efficiency and informativeness are very good across the farms and present little variability with prediction sets having an average size of 1 to 2 classes and around $80\%$ of prediction sets being singletons. In some cases, such as \textit{Farm 7}, the model presents exceptionally good performance. The coverage is maintained exactly at the required level, with very low average set size and $90\%$ of predicted sets being \textit{weed} singletons. This is probably due to the fact that the images obtained in this farm show clearly the distinction between the weeds and the cultivated crops, with a nice contrast between the plants and the ground, as can be seen in Figure \ref{fig:example_images}. Under such conditions, the model does not exhibit high uncertainty in its predictions.

\begin{table}[h]
\centering
\begin{tabular}{rccc}
\multicolumn{1}{c}{\textbf{Farm}}                       & \textbf{Coverage}    & \textbf{Efficiency}  & \textbf{Informativeness} \\ \hline
\multicolumn{1}{r|}{\textit{(New) Farm 0}}              & 0.94                 & 1.13                 & 0.88                     \\ \hline
\multicolumn{1}{r|}{\textit{(New) Farm 4}}              & 0.88                 & 1.24                 & 0.79                     \\ \hline
\multicolumn{1}{r|}{\textit{(New) Farm 5}}              & 0.94                 & 1.38                 & 0.67                     \\ \hline
\multicolumn{1}{r|}{\textit{(New) Farm 7}}              & 0.90                 & 1.11                 & 0.90                     \\ \hline
\multicolumn{1}{r|}{\textit{(New) Farm 8}}              & 0.94                 & 1.23                 & 0.79                     \\ \hline
\multicolumn{1}{r|}{\textit{(New) Farm 10}}             & 0.92                 & 1.20                 & 0.82                     \\ \hline
\multicolumn{1}{r|}{\textit{(New) Farm 20}}             & 0.78                 & 1.35                 & 0.71                     \\ \hline
\multicolumn{1}{r|}{\textit{(New) Farm 21}}             & 0.94                 & 1.19                 & 0.83                     \\ \hline
\multicolumn{1}{r|}{\textit{(New) Farm 23}}             & 0.85                 & 1.39                 & 0.66                     \\ \hline
\multicolumn{1}{r|}{\textit{(New) Farm 26}}             & 0.90                 & 1.27                 & 0.76                     \\ \hline
\multicolumn{1}{r|}{\multirow{2}{*}{\textbf{Marginal}}} & \textbf{0.90}        & \textbf{1.24}        & \textbf{0.79}            \\
\multicolumn{1}{r|}{}                                   & \multicolumn{1}{c}{$(\pm 0.05)$} & \multicolumn{1}{c}{$(\pm 0.10)$} & \multicolumn{1}{c}{$(\pm 0.08)$}    
\end{tabular}
\caption{Conformal prediction results on class \textit{weed} for each farm in the \textit{new} database using the PIP nonconformity score (standard deviation between parenthesis).}
\label{tab:exp2-conf-results}
\end{table}

\begin{figure}[h]
    \centering
    \includegraphics[width=0.5\linewidth]{ 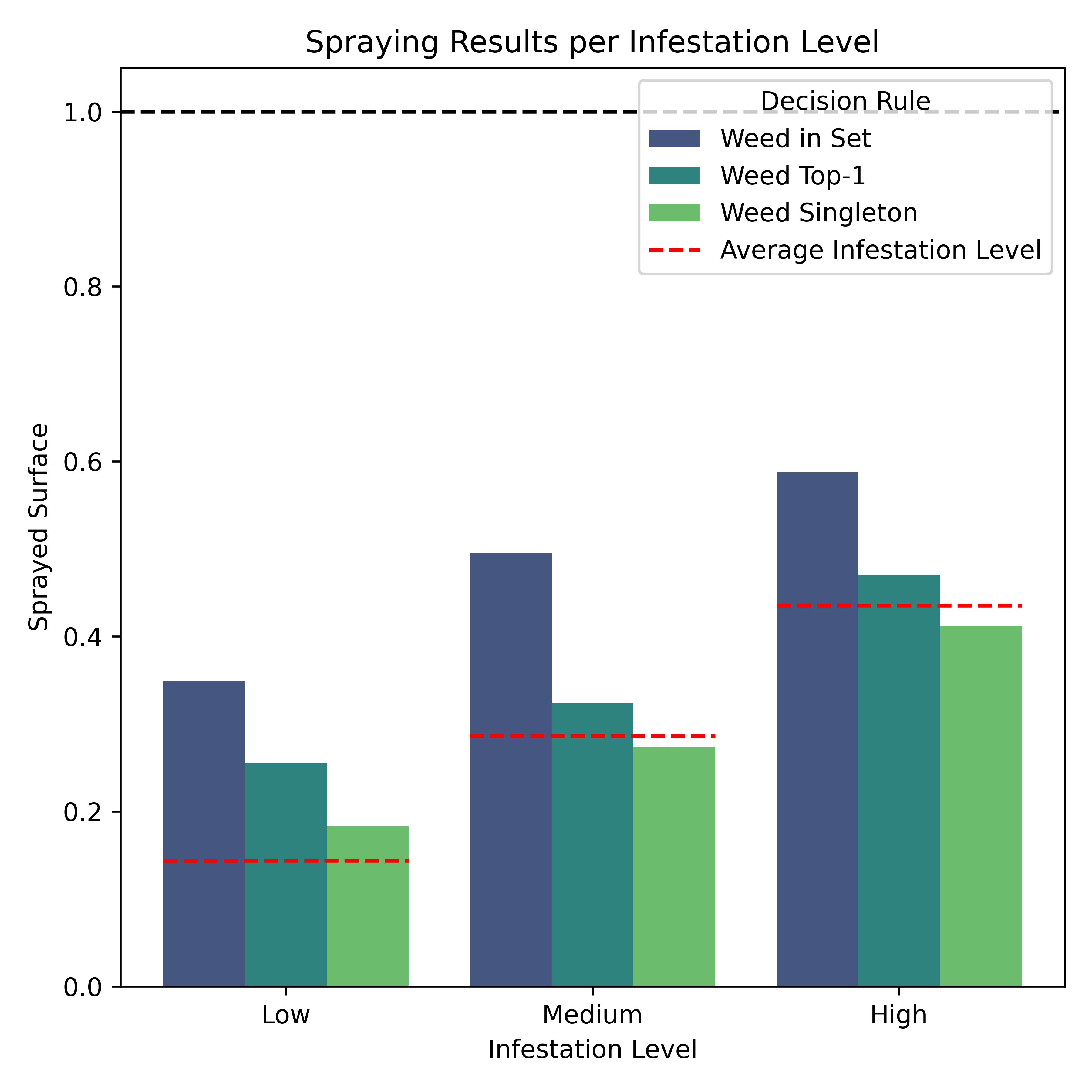}
    \caption{Bar plot of spraying results per level of infestation. \textit{Low} infestation corresponds to less than $20\%$ of infestation, \textit{Medium} to around $20-40\%$ and \textit{High} to more than $40\%$ infestation. The black dashed line represents broadcast spraying ($100\%$ of the surface is sprayed), and the red dashed lines represent the average infestation level in each category. The Spray Surplus and Spray Reduction can be read by comparing the bars to the red and black dashed lines, respectively.}
    \label{fig:exp2_barplot}
\end{figure}

In order to study the performance of the three spraying decision rules over the different farms, we categorize the farms by their Infestation Level. Three categories are defined: \textit{Low} corresponds to a level of infestation reaching $20\%$ of total farm surface, \textit{Medium} corresponds to levels between $20$ and $40\%$ and \textit{High} for levels of infestation superior to $40\%$. Figure \ref{fig:exp2_barplot} visualizes all the spraying results for three infestation categories. Each bar corresponds to the Sprayed Ratio induced by each of the decision rules. For each infestation category, the average level of infestation is shown by the red dashed line. The difference between the sprayed ratio and the red dashed line represents the average spray surplus. On the other hand, Spray Reduction can be read by comparing the black dashed line -- representing broadcast spraying that covers 100\% of the surface -- with the bar plots for each decision rule.

In line with the results of the previous experiment, \textit{Weed in Set} is the one that manifests the highest level of surplus. This is due to it being the least stringent decision rule, as stated previously. \textit{Weed Singleton} is such a constraining rule that it fails to cover all the weeds for the farms manifesting \textit{Medium} and \textit{High} levels of infestation. \textit{Weed Top-1} manifests a balanced behavior in terms of Spray Surplus. However, \textit{Weed in Set} remains the only decision that formally guarantees covering the weeds at the required level, even if it over-sprays. Indeed, the fact that the two other decision rules lead to a sprayed ratio that is not far away from the infestation level is a purely empirical observation that cannot be formally guaranteed. In addition, these decision rules provide no guarantee on the \textit{weed} coverage. This means that even though the spray ratio is close to the infestation level, we cannot be sure that the spraying decisions were taken at the right locations.

\begin{table}[h]
\resizebox{\textwidth}{!}{
\begin{tabular}{c|cc|cccc}
\hline
\textbf{Farm}          & \textbf{Recall}  & \textbf{Precision} & \textbf{Infestation Level} & \textbf{Sprayed Ratio} & \textbf{Spray Reduction} & \textbf{Spray Surplus} \\ \hline
\textit{(New) Farm 0}  & 0.94             & 0.65                & 0.42                        & 0.60                    & 0.40                    & 0.18                   \\ \hline
\textit{(New) Farm 4}  & 0.88             & 0.73                & 0.36                        & 0.43                    & 0.57                    & 0.07                   \\ \hline
\textit{(New) Farm 5}  & 0.94             & 0.32                & 0.24                        & 0.69                    & 0.31                    & 0.45                   \\ \hline
\textit{(New) Farm 7}  & 0.90             & 0.43                & 0.22                        & 0.46                    & 0.54                    & 0.24                   \\ \hline
\textit{(New) Farm 8}  & 0.94             & 0.60                & 0.37                        & 0.57                    & 0.43                    & 0.20                   \\ \hline
\textit{(New) Farm 10} & 0.92             & 0.62                & 0.27                        & 0.39                    & 0.61                    & 0.13                   \\ \hline
\textit{(New) Farm 20} & 0.78             & 0.34                & 0.17                        & 0.39                    & 0.61                    & 0.22                   \\ \hline
\textit{(New) Farm 21} & 0.94             & 0.36                & 0.12                        & 0.30                    & 0.70                    & 0.18                   \\ \hline
\textit{(New) Farm 23} & 0.85             & 0.53                & 0.27                        & 0.43                    & 0.57                    & 0.16                   \\ \hline
\textit{(New) Farm 26} & 0.90             & 0.71                & 0.46                        & 0.58                    & 0.42                    & 0.12                   \\ \hline
\textbf{Marginal}      & \textbf{0.89}    & \textbf{0.53}       & \textbf{0.45}               & \textbf{0.50}           & \textbf{0.50}           & \textbf{0.20}           \\
\end{tabular}
}
\caption{Spraying results for each farm using the \textit{Weed in Set} decision rule.}
 \label{tab:exp2-spray-results}
\end{table}

Table \ref{tab:exp2-spray-results} details the spraying metrics using the \textit{Weed in Set} decision rule. The recall (coverage) is maintained for all farms with some variability, mostly evident for the farms that have shown inferior conformal results (\textit{Farm 20} and \textit{Farm 23} for example). The spraying precision exhibits much higher inter-farm variability and is, on average, quite low (around 53\%), which should signal a relatively high level of spray surplus. Indeed, the average spray surplus is around 20\% of the total surface, although for some farms it is as low as 7\% while for others (\textit{Farm 5}, which has the lowest precision) it is around 45\%. In all cases, this spraying pipeline still demonstrates significant reduction in spraying amount with respect to full broadcast spraying. This reduction ranges from 31\% (\textit{Farm 5}) for the farm where the model is ``struggling'' the most, to up 70\% (\textit{Farm 21}) with an average spray reduction of 50\%.

\section{Discussion} \label{sec:discussion}
As deep learning-based precision spraying starts to get deployed in the real-world, it becomes increasingly important to improve the reliability of the predictive systems that provide the backbone of these spraying pipelines. As Darbyshire \textit{et al.} \citep{darbyshireReviewWeedRecognition2024} state in their recent article: ``improvements in the reliability and accuracy of weed
recognition are crucial for the successful implementation of precision targeted weed management systems.'' It is important to note, however, that the focal point of our work is not the maximization of the predictive performance of the deep learning models, as is the case with the vast majority of recent works in the precision weeding literature \citep{raiApplicationsDeepLearning2023, hasanSurveyDeepLearning2021}. Rather, this article tackles a deeper, longer-term, problem that is the development of reliable, trustworthy and certifiable predictive models. Although the construction of larger and more comprehensive datasets and the development of new architectures and learning techniques can play a role in improving the reliability of the systems \citep{raiApplicationsDeepLearning2023, darbyshireReviewWeedRecognition2024, hasanSurveyDeepLearning2021}, they are not enough to provide valid distribution-free performance guarantees. The current research provides a starting point for the development, improvement and integration of such methodologies as conformal prediction in the ag-tech ecosystem, which lags behind other domains on this front. 

Furthermore, the current work takes a broader perspective than simply focusing on the deep learning models. It considers the full spraying pipeline in accordance with Darbyshire \textit{et al.}'s \citep{darbyshireReviewWeedRecognition2024} recommendation that ``optimizing [machine learning] approaches without considering the systems into which they will be integrated may be short-sighted.'' Indeed, the current work evaluates the predictive pipeline at multiple levels: the standard classification, the conformal prediction, and the spraying decision levels, while making sure that the data acquisition and model development lifecycles resemble real-world scenarios. That is, the experimental setup considered in this article was developed to be somewhere between the ``in lab'' and real-world settings. Certain assumptions taken are rarely satisfied under real-world deployment. For example, the object of interest (crop or weed) is often observed multiple times as the system advances in the parcel -- a fact that was not taken into consideration in our experimental setup. In addition, chemical spray drift often occurs due to wind, which means that each image cannot be mapped one-to-one to a spraying outcome. However, these assumptions are crucial to permit the experimental study of the current methodology in the absence of field testing. Future works may involve a relaxation of these assumptions and accompanied by a field evaluation of the proposed pipeline. 

The experimental results have shown the validity of the conformal prediction approach in guaranteeing the coverage of weeds both \textit{in-distribution} (Experiment 1), and under conditions of ``shifted'' or ``near'' \textit{out-of-distribution} \citep{winkensContrastiveTrainingImproved2020, mukhotiRaisingBarEvaluation2023} (Experiment 2), both scenarios that naturally occur when deploying precision weeding systems in the real-world. Without re-training and calibrating the neural network, a conformal predictor can be deployed in a new farm and will maintain the coverage as long as the prediction domain is not ``far away'' from the data on which the model has been calibrated. Marginally over all deployment domains, the system is guaranteed to maintain the $1-\alpha$ coverage level. While the current work presented only the marginal and class-conditional split conformal approaches, certain extensions such as group-conditional \citep{angelopoulosConformalPredictionGentle2023a, melkiGroupConditionalConformalPrediction2023c} and clustered \citep{dingClassConditionalConformalPrediction2023} conformal prediction can be used to guarantee the required coverage for any sub-group of examples in the population as defined by auxiliary criteria (for example, the \textit{farm} or other environmental characteristics). As such, the conformal methodology can be adapted to the context at hand.

The versatility of the conformal approach is also due to the fact that it allows the model to produce prediction sets, rather than simple point predictions. These predictions sets are \textit{adaptive} in the sense that their size reflects the base model's uncertainty about the observed example. This also offers the practitioner the ability to define any decision rule on these prediction sets. In the current work, three basic decision rules have been studied: \textit{Weed in Set}, \textit{Weed Top-1} and \textit{Weed Singleton}. While \textit{Weed in Set} is the only decision rule that guarantees the spraying coverage at $1-\alpha$, the user is still free to choose other rules. The choice depends on the importance given to precision, which is related to the tolerance to spray surplus, and more generally to the user's preference between exact weed coverage and spray reduction. This opens the door to the deployment of systems that are more modular and adaptive to users' requirements.

\section{Conclusion} \label{sec:conclusion}
This article constitutes a first comprehensive study, to the authors' knowledge, of the utility of conformal prediction in precision agriculture, and particularly in precision weeding. We have presented, in detail, the conformal prediction methodology and its main components. Particular focus was accorded to the split-conformal algorithm that can be applied marginally to guarantee a $1-\alpha$ coverage level on average over all observations, or on a class-conditional basis to guarantee the coverage level for each class -- and \textit{a fortiori} on the class of interest \textit{weed}. This methodology is an important step towards the development of valid, trustworthy and certifiable deep learning-based autonomous systems in agriculture that can earn the trust of the farmers and thus increase the adoption of these technologies \citep{daraRecommendationsEthicalResponsible2022, alexanderWhoResponsibleResponsible2023}.

Conformal prediction was studied in the context of a fully automated precision spraying pipeline based on image classification. Three spraying decision rules were defined on top of the predicted conformal sets to lead to a final binary \texttt{spray / no spray} decision. This pipeline was evaluated at three levels: (1) the level of standard classification, using common metrics such as precision, recall and the f1-score, (2) the level of conformal classification using the empirical coverage -- which was shown to be maintained at the $1-\alpha$ level, and the efficiency and informativeness of the prediction sets, (3) and finally at the spraying decision level. 

Two experimental procedures that reflect commonly faced scenarios were described in detail. The first experiment demonstrates the power of the conformal approach when the prediction domain is similar to the data seen during conformal calibration. The spraying results show that we can guarantee covering at least $1-\alpha = 90\%$ of the weeds while reducing the sprayed quantity by $60\%$ with respect to broadcast spraying. The second experimental setup shows how conformal prediction allows the predictive model to maintain its weed coverage guarantee on new farms that may not always be aligned with the previously seen data, and that, without re-training the neural network. Such a result is especially useful for autonomous systems that will be deployed in new, uncontrolled, and unobserved environments. It improves the robustness and reliability of the systems while providing signals (through worse coverage, efficiency and informativeness) when the model is not performing well. 

It is the authors' hope that this work will motivate the ag-tech research community to start exploring conformal prediction, and more largely, the domain of deep learning safety, with the aim of developing systems that can be trusted and certified.

\section*{Declarations}

\noindent \textbf{Acknowledgments} \quad The authors would like to thank Estelle Millet and the AgroData team at \textit{EXXACT Robotics} for the acquisition, annotation and management of the data used in this article; and Vincent Rachet, CEO of \textit{EXXACT Robotics}, for his valuable input towards the finalization of this article. \\

\noindent \textbf{CRediT statement} \quad \textbf{Paul Melki:} Conceptualization, Methodology, Software, Validation, Formal Analysis, Investigation, Data Curation, Writing - Original Draft, Writing -- Review \& Editing, Visualization; \textbf{Lionel Bombrun: } Conceptualization, Methodology, Formal Analysis, Validation, Writing -- Review \& Editing, Supervision; \textbf{Boubacar Diallo: } Conceptualization, Methodology, Software, Resources, Data Curation, Writing - Review \& Editing, Supervision; \textbf{Jérôme Dias: } Conceptualization, Methodology, Validation, Resources, Writing -- Review \& Editing, Supervision, Project Administration, Funding Acquisition; \textbf{Jean-Pierre Da Costa:} Conceptualization, Methodology, Formal Analysis, Validation, Writing -- Review \& Editing, Supervision, Project Administration, Funding Acquisition
\\

\noindent \textbf{Funding} \quad Paul Melki’s PhD thesis is funded by the \textit{Association Nationale de Recherche et Technologie} (ANRT, France) under the CIFRE contract 2021/1200. \\

\noindent \textbf{Conflict of interest} \quad The authors declare that they have no known competing financial interests or personal relationships that could have influenced the work reported in this paper. \\

\noindent \textbf{Data availability} \quad Due to confidentiality reasons, the data used in this study cannot be made public at the moment, but may be released at a later stage.

{\small
\bibliographystyle{apalike}
\bibliography{biblio}
}

\end{document}